# RTAT: A Robust Two-stage Association Tracker for Multi-Object Tracking

Song Guo[1], Rujie Liu[1] and Narishige Abe[2]

[1] Fujitsu Research and Development Center Co., Ltd., Beijing, China
[2] Fujitsu Ltd., Kanagawa, Japan
guosong@fujitsu.com

**Abstract.** Data association is an essential part in the tracking-by-detection based Multi-Object Tracking (MOT). Most trackers focus on how to design a better data association strategy to improve the tracking performance. The rule-based handcrafted association methods are simple and highly efficient but lack generalization capability to deal with complex scenes. While the learnt association methods can learn high-order contextual information to deal with complex scenes, but they have the limitations of higher complexity and cost. To address these limitations, we propose a Robust Two-stage Association Tracker, named RTAT. The first-stage association is performed between tracklets and detections to generate tracklets with high purity, and the second-stage association is performed between tracklets to form complete trajectories. For the first-stage association, we use a simple data association strategy to generate tracklets with high purity by setting a low threshold for the matching cost in the assignment process. We conduct the tracklet association in the second-stage based on the framework of message-passing GNN. Our method models the tracklet association as a series of edge classification problem in hierarchical graphs, which can recursively merge short tracklets into longer ones. Our tracker RTAT ranks first on the test set of MOT17 and MOT20 benchmarks in most of the main MOT metrics: HOTA, IDF1, and AssA. We achieve 67.2 HOTA, 84.7 IDF1, and 69.7 AssA on MOT17, and 66.2 HOTA, 82.5 IDF1, and 68.2 AssA on MOT20.

## 1 Introduction

Multi-Object Tracking (MOT) aims to detect and identify all the objects, and ideally to form one complete trajectory for each object in a video. It is an essential technology for various applications, such as intelligent surveillance, autonomous driving, and robotics. Tracking-by-Detection (TbD) [1, 3, 4, 7] is currently the most effective paradigm for MOT, which contains two steps: object detection and data association. Most trackers [1, 4, 5, 7, 8, 10] focus on how to design a better data association strategy to enhance the tracking performance. Various strategies have been proposed and they broadly fall into two categories: handcrafted association and learnt association.

Generally, the handcrafted association methods first compute the distances between tracklets and detections, and then match them according to their distances [1, 4, 5, 7, 8, 10, 33, 34]. The matching is usually done by bipartite matching, which is formulated as a Linear Assignment Problem (LAP) and solved by Hungarian algorithm [27]. These methods explicitly leverage various cues (e.g., location, motion, appearance) to calculate the distances, and then design various strategies to construct the cost matrix to perform identity assignment. Because of their simplicity and efficiency, these methods are very popular in MOT. However, most of them are rule-based, so it is hard and exhausting to design a generic association strategy that can deal with various scenes, such as crowded, fast camera motion, night, and low resolution. Another drawback of these methods is that the association error cannot be fixed once it occurs.

In the learnt association methods, the data association is usually performed implicitly based on the learnt association feature through a neural network, such as Transformer [28, 29, 30], Graph Neural Networks (GNN) [17, 18, 20, 21, 22]. These methods learn to extract high-order association feature from multiple sources of information (e.g., spatial and temporal, appearance, motion, and etc.) in a data-driven manner, where the matching is implicitly performed without using bipartite matching. Transformer-based methods [28, 30] perform data association by query propagation. However, the training strategies are highly complex and it needs a large amount of data to train Transformer models, while the scales of MOT datasets are often limited [25, 26]. Graph-based methods [18, 21, 22] first construct graphs in which nodes represent the detections and edges indicate their connections, and the data association is treated as an edge classification problem. However, it needs to construct very large graphs for long videos or videos in crowded scenes, which brings the issues of high computational complexity and large memory consumption [21]. Generally speaking, the learnt association methods can leverage high-order information to deal with more complex scenes, but they have the limitations of higher complexity and cost.

In order to effectively utilize the advantages and address the limitations of these two kinds of data association methods, we propose a Robust Two-stage Association Tracker, named RTAT. The first-stage association is performed between tracklets and detections to generate tracklets with high purity, and the second-stage association is performed between tracklets to form complete trajectories.

In the first-stage association, we use a simple data association strategy to generate tracklets with high purity. This is done by setting a low threshold for the matching cost in the identity assignment process. The generated tracklets have higher purity and less identity switches. As a result, the number of tracklets will increase, and the fragmentations problem will be solved in the second-stage by using tracklets association.

In the second-stage association, we merge the tracklets into complete trajectories by using GNN. Our method models the tracklet merging as a series of edge classification problem in hierarchical graphs, which can recursively merge short tracklets into longer ones and finally form complete trajectories. We use the message passing mechanism [31, 21] to update the graphs and learn features for nodes and edges, and then perform edge classification based on the final edge feature. This process is hierarchically performed on graph in each level. Since the number of tracklets is much smaller than that of detections, our GNN model takes all the tracklets in a video se-quence as input. Experiments on two of the most popular MOT benchmarks: MOT17 [25] and MOT20 [26], demonstrate the effectiveness of our method.

## 2 Related works

### 2.1 Handcrafted association

The handcrafted association methods match the detections to the tracklets based on well-designed cost matrix by leveraging various strategies [1, 4, 5, 7, 8, 10, 33, 34]. Intersection over union (IoU) and appearance distance are the most commonly used metrics to construct the cost matrix. Motion model is adopted to predict the locations of tracklets to calculate the IoU distance with the detections, while person Re-identification (ReID) model is used to extract the appearance features to calculate the appearance distance. Generally, IoU distance is more useful in short-term matching, while appearance information is more accurate in long-term matching.

There are four main research directions in the handcrafted association methods. (1) Learn more accurate motion models: Kalman filter (KF) and its variants [32, 33, 34, 4, 1, 5], camera motion compensation (CMC) [4, 5], etc. (2) Extract more discriminative ReID feature: independent ReID model [34, 4, 7], occlusion-aware ReID feature [35], dynamic ReID feature [4, 5], etc. (3) Design more sophisticated strategy to construct the cost matrix: different combination of the IoU and appearance distance, such as weighted sum [5, 7, 36], minimum cost [4], etc. (4) Develop better matching strategy: single matching [33], cascade matching [34, 2, 3], etc. Many researchers have invested a great deal of time and effort in designing a better data association strategy. However, it is hard and exhausting to design a generic data association that can deal with various scenes. Therefore, we turn to use a simple association method to obtain tracklets with high purity, and further merge them by using tracklet association.

### 2.2 Graph-based association

Graph-based methods perform data association on constructed graphs, where nodes represent detections and edges indicate linkage between them. The data association is formulated as a graph optimization problem, which is solved by different algorithms, such as network flows [15], k-shortest paths [16], minimum cost lifted multicut [17], lifted disjoint paths [18, 19], etc. Recently, GNN [20] is introduced as an extension of neural networks that can operate on graph. GNN can extract high-order contextual information by adopting a message passing mechanism, which propagates the information encoded in the features of neighboring nodes and edges across the graph [20, 21, 22, 23]. MPNTrack [21] designs a tracker based on Message Passing Network to learn features for nodes and edges and treats the data association as an edge classification task. SUSHI [22] proposes a unified tracker for short and long-term tracking by using a hierarchy of message passing GNNs. SGT [23] employs GNNs to recover the missed detections to enhance the tracking performance for online graph tracker.

In contrast to handcrafted association, graph-based association methods seek for global optimization over longer range frames. Specially, GNN-based methods can learn high-order information through message passing, and therefore they can achieve better tracking performance [21, 22]. However, it needs to construct very large graphs for long videos or videos in crowded scenes, which brings the issues of higher complexity and cost [21]. In our work, we build graph for tracklet association, where the scale of the graph is much smaller. Therefore, it can effectively solve the above-mentioned problems and still utilize the advantages of graph-based association.

### 2.3 Tracklet association

Tracklet association [11, 12, 13] has drawn much attention in TbD based MOT. Several methods [11] exploit the idea of multi-level association, which first generates short tracklets in adjacent frames and then merges them into trajectories by tracklet association. Some works [12, 13] follow the split-merge pipeline to refine the tracking results of existing trackers, and tracklet association is employed in the merging process. TAT [11] employs a Multi-Layer Perceptron (MLP) to link detections in adjacent frames to generate short tracklets, and then trains a network flow to associate the tracklets into trajectories. ReMOT [12] splits tracklets by using appearance and motion features, and then associates the tracklets by hierarchical clustering on a designed distance matrix. [13] proposes a tracklet booster for existing trackers, which trains a Splitter to split tracklets into small pieces, and then learns a Connector to merge the tracklet pieces that are from the same identity. These methods generate short tracklets either in a sliding window with limited size or by splitting existing tracklets into small pieces. The generated tracklets are often too short, which will increase the burden for the following tracklet association. Furthermore, performing tracklet association by using the message-passing GNN has not been fully exploited in these methods.

## 3 Methodology

### 3.1 Motivation

The motivation of our Robust Two-stage Association Tracker (RTAT) is simple and effective. It is hard and exhausting to design a generic data association strategy that can handle various scenes by explicitly leveraging simple cues, while learnt association methods have the limitations of higher complexity and cost, although they can learn high-order information to deal with more complex scenes. Therefore, we propose to use simple cues to generate clean tracklet pieces, and then employ GNN for tracklet association to obtain the final trajectories. RTAT consists of two-stage associations, where the first-stage association is performed between tracklets and detections to generate tracklets with high purity, and the second-stage association is performed between tracklets to obtain complete trajectories. The workflow of RTAT is shown in Fig. 1. We will describe the details of our method in the following sections.

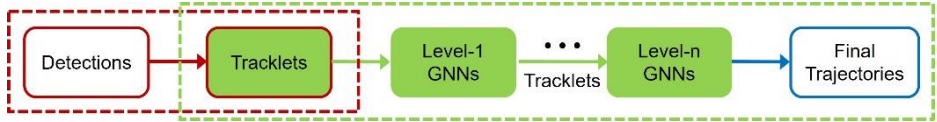

**Fig. 1.** The workflow of our robust two-stage association tracker (RTAT). The first-stage association (red dashed box) generate tracklets with high purity from detections, and the second-stage association (green dashed box) merge short tracklets into longer ones by using hierarchical GNNs and finally form complete trajectories

### 3.2 Method formulation

Given a video sequence with $K$ frames and a set of detections $D = \{d_i, i \in [1, M]\}$, where $M$ is the total number of detections obtained from the $K$ frames. Each detection $d_i$ can be represented by its bounding box coordinates, image region, and timestamp. Let us define the set of tracklets as $T = \{t_j, j \in [1, N]\}$, where $N$ is the number of tracklets in the video sequence. Each tracklet consists of a set of detections $t_j = \{d_j^i, i \in [1, n_j]\}$, where $n_j$ is the number of detections in the tracklet of $t_j$. The aim of our first-stage association is to generate the initial set of tracklets $T$.

In the task of tracklet association, we construct an undirected graph $G = (V, E)$, where nodes represent the tracklets (e.g., $V = T$) and edges indicate the connections between them. The set of edges can be denoted as $E = \{e_{ij} = (t_i, t_j) \in N \times N, i \neq j\}$, where $e_{ij}$ represents the linkage of a pair of tracklets $(t_i, t_j)$. We introduce a binary variable $y_{e_{ij}}$ to indicate whether $t_i$ and $t_j$ are from the same identity. Specifically, if they are from the same identity $y_{e_{ij}} = 1$ and the edge $e_{ij}$ is active, otherwise it is inactive. We perform edge classification to predict the values of each edge based on the learnt edge feature and merge the tracklets belong to the same identity, i.e., nodes are linked by active edge. Different from other graph-based association methods which take detections in a short video clip with limited number of frames as input, we take all the tracklets in a video as input to obtain the final trajectories.

### 3.3 First-stage: tracklet generation

The aim of the first-stage association is to generate tracklets with high purity. Any tracker can be used in this stage, but we prefer trackers with simple data association strategy, such as ByteTrack [1], BoT-SORT [4]. The matching is usually done by bipartite matching, which is solved by Hungarian algorithm [27]. In the assignment process, we set a cost threshold $th_c$ for possible matching and reject the matchings with higher cost than $th_c$. For simplification, we normalize the value of the cost in cost matrix to be [0, 1] for different tracker. By setting a lower cost threshold $th_c$, we can obtain tracklets with higher purity. Consequently, there are less identity switches in each tracklet, but the number of tracklet fragments will increase. We will focus on solving the fragmentation problem in the next stage by using tracklets association.

### 3.4 Second-stage: tracklet association

The aim of the second-stage association is to merge the tracklet pieces into trajectories. We perform the tracklet association based on the framework of message-passing GNN [31, 21, 22]. An illustration of the tracklets merging process is shown in Fig. 2. Our method models the tracklet merging as a series of edge classification problem in hierarchical graphs, which can recursively merge short tracklets into longer ones. We use the message passing mechanism to update the feature vectors for nodes and edges across the graph and the edge classification is performed based on the final edge feature. This process is performed hierarchically for graph in each level and the workflow of each level contains four main steps:

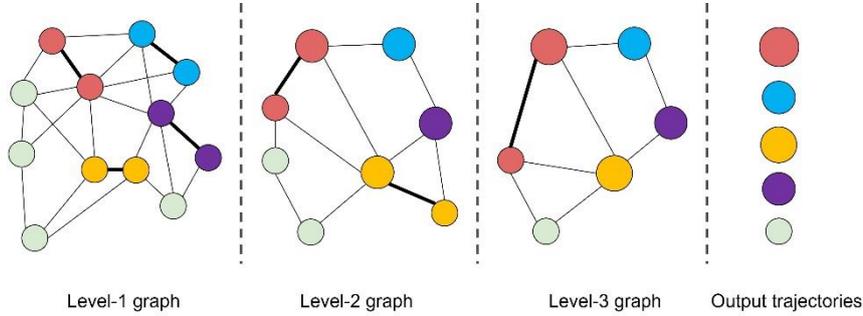

**Fig. 2.** An illustration of the tracklets merging process in hierarchical graphs. The bold edges are classified as active, and the nodes linked by active edge will be merged in current level. The final trajectories are listed in the last column.

**Graph construction.** We construct an undirected graph $G = (V, E)$, where each node represents a tracklet and each edge indicates the possible connection for a pair of tracklets. Compared to detection association, the number of nodes is largely reduced for tracklet association. However, the number of edges is still very large if all the connections between each pair of nodes are considered. Moreover, it will cause a severe label imbalance between active and inactive edges. Therefore, we only consider the edge between a pair of nodes that have no temporal overlap. We further limit the number of edges for each node to be $K$, which are selected by its top $K$ nearest neighbors according to the similarity measures of appearance, motion, and spatial position. Hence, we construct a sparse graph with limited number of edges, which can reduce the computational complexity, and alleviate the edge label imbalance problem.

**Graph initialization.** The node feature vector is initialized by the feature of its corresponding tracklet. We first extract appearance features for all the detections in each tracklet, and then calculate their average feature as the tracklet feature. The averaged tracklet feature is more robust to motion blur, partial occlusion, and illumination change than single detection appearance feature. The tracklet feature is fed into a node encoder $E_n^{enc}$, whose output is used to initialize its corresponding node feature.

The edge feature vector is initialized with the output of an MLP, the input of which is a concatenated vector of the association features from two connected tracklets. We adopt spatial and temporal distance, appearance and motion information to construct the initial feature vector, which is an extension of MPNTrack [21] and SUSHI [22].

For a pair of tracklets $T_a$ and $T_b$ with their detection box coordinates and timestamps, which can be described as $T_a = \{(x_i, y_i, w_i, h_i, t_i), i \in [a_1, a_n]\}$ and $T_b = \{(x_j, y_j, w_j, h_j, t_j), j \in [b_1, b_n]\}$, where $[a_1, a_n]$ and $[b_1, b_n]$ are the frame range of $T_a$ and $T_b$ respectively. Assuming that $T_a$ ends before $T_b$ starts, so we have $t_{a_n} < t_{b_1}$. We use their closest detection boxes to compute the relative spatial distance and scale difference, which is formulated as:

$$\left[\frac{2(x_j-x_i)}{h_j+h_i}, \frac{2(y_j-y_i)}{h_j+h_i}, \log\frac{h_j}{h_i}, \log\frac{w_j}{w_i}\right] \tag{1}$$

where $i = a_n$ and $j = b_1$. Supposing the FPS (Frames Per Second) of the given video is $fps$, we calculate their time difference by the formula: $(t_{b_1} - t_{a_n})/fps$.

To encode the appearance information, we use the Euclidean distance of the tracklets feature and the average cosine similarity of the top $L$ closest detections for each pair of tracklets, which can be formulated as:

$$\left[\left\|app_{T_b}^{avg} - app_{T_a}^{avg}\right\|_2, \frac{1}{L*L}\sum_{i=1}^{L}\sum_{j=1}^{L} \cos(app_{T_a}^i, app_{T_b}^j)\right] \tag{2}$$

where the first distance encodes global appearance discrepancy between two tracklets, and the second similarity describes local appearance similarity, which is helpful to remove the influence of large appearance variations inside a tracklet, such as large pose, long-time occlusion, and etc.

The tracklets belong to the same trajectory are expected to satisfy motion consistency [22], so we add the motion information into the edge feature. For a pair of tracklets $T_a$ and $T_b$, we calculate their middle frame $t_{mid} = t_{a_n} + (t_{b_1} - t_{a_n})/2$, and predict their box positions at this frame using KF, which are respectively denoted as $pred\_box_{T_a}^{t_{mid}}$ and $pred\_box_{T_b}^{t_{mid}}$. We adopt the Generalized Intersection over Union (GIOU) [6] score of these two estimated boxes to measure their motion consistency: $GIOU(pred\_box_{T_a}^{t_{mid}}, pred\_box_{T_b}^{t_{mid}})$

Finally, the concatenation of these feature vectors, the dimension of which is 8, is fed into an edge encoder $E_e^{enc}$ to obtain the initial edge feature.

**Graph update.** We employ the message-passing mechanism to update the features for nodes and edges [31, 21, 22]. During each step of message-passing, every node and edge aggregates their received information, and then combine the incoming information with their own to update their feature vectors [31]. Specifically, for the construct graph $G = (V, E)$, we obtain the initial feature vector $f_i^0$ and $f_{(i,j)}^0$ for each node $i \in V$ and each edge $(i, j) \in E$ from the graph initialization step. The mechanism of message-passing is to propagate messages between neighboring nodes and edges across the graph. The propagation is performed by alternately updating the features of edges and nodes, which is divided into two steps: update edge feature using neighboring nodes and update node feature using neighboring edges. Both updates are sequentially performed for $L$ iterations. For each iteration $l \in [1, L]$, the edges and nodes features are updated as follows:

$$f_{(i,j)}^l = U_e([f_i^{l-1}, f_j^{l-1}, f_{(i,j)}^{l-1}]), \; m_{(i,j)}^l = U_n([f_i^{l-1}, f_{(i,j)}^l]), \; f_i^l = \phi\left(\{m_{(i,j)}^l\}_{j \in N_i}\right) \quad (3)$$

where $U_e$ and $U_n$ are MLP networks that aggregate information from neighboring nodes and edges. $N_i$ is the set of nodes adjacent to node $i$, and $\phi$ denotes an order-invariant operation, e.g., maximum, summation or average. After $L$ iterations, we obtain the final node and edge features, which contain high-order contextual information from neighboring nodes and edges in a distance of $L$ along the graph.

**Edge classification.** We use an MLP with sigmoid function as the edge classifier $C_e^{class}$ and then perform edges classification based on their final features $f_{(i,j)}^L$.

$$y_{(i,j)} = C_e^{class}(f_{(i,j)}^L), (i,j) \in E \quad (4)$$

where the predicted edge score $y_{(i,j)} \in (0, 1)$. The scores are further rounded to binary values using the exact rounding solution described in [21]. The edges are classified as active or inactive, and the tracklets linked by the active edges are merged into longer ones. We update the set of tracklets and hierarchically perform the four steps, i.e., graph construction, graph initialization, graph update and edge classification.

**Data augmentation.** In this stage, a training sample consists of a video sequence and a set of tracklets. There are very few training samples in MOT17 [25] and MOT20 [26], which are 7 and 4 respectively. Therefore, we introduce data augmentations from both video-level and tracklet-level to train the GNN networks with higher robustness and generality. In video-level augmentation, we generate more video clips from the original video sequences. We sample a video clip in every 50 frames (i.e., start points), and the start frame is randomly selected with a fluctuation of 15 frames at each start point. The length of a video clip is randomly selected from 25% to 100% of the length for the whole video. In tracklet-level augmentation, we generate more sets of tracklets by adopting different data association strategies under different cost thresholds in Section 3.3.

**Training GNN.** We use the same GNN architectures for graphs in different hierarchical levels. Since the aims of all hierarchical levels are the same, which is to merge tracklets that belong to the same identity into longer ones, we also share the parameters for all hierarchical levels. The difference among different levels is the lengths and numbers of tracklets, so we learn a level adapter which is added to edge feature in each level of GNN. The level adapter will help the GNN model to learn the most important cues for each level in a data-driven manner. We adopt the focal loss to train the edge classifier in each level and the loss is a summation of the losses in all levels.

## 4 Experiments

### 4.1 Experimental Settings

**Datasets.** We conduct our experiments on two of the most popular MOT benchmarks: MOT17 [25] and MOT20 [26], under the "private detection" protocol. MOT17 [25] contains 14 video sequences which are filmed under a variety of conditions, such as different camera motions, viewpoints, and weather conditions. MOT20 [26] contains 8 video sequences in very crowded scenes.

**Metrics.** Our method focuses on robust data association, so we adopt HOTA [24] and IDF1 [39] as the main metrics. We also use the metrics MOTA, AssA [24], and IDs to provide comparisons from more perspectives. HOTA maintains a good balance between the accuracy of object detection and association. IDF1 measures the identity preservation ability and focus more on the association ability. AssA is used to evaluate the association performance, while MOTA focuses on the detection performance.

We introduce a new metric, named High Purity Rate (HPR), to measure the rate of high purity tracklets in a video sequence. A tracklet has high purity if more than 80% of its detections are from the same identity. HPR is the rate of high purity tracklets in all the tracklets for a given video.

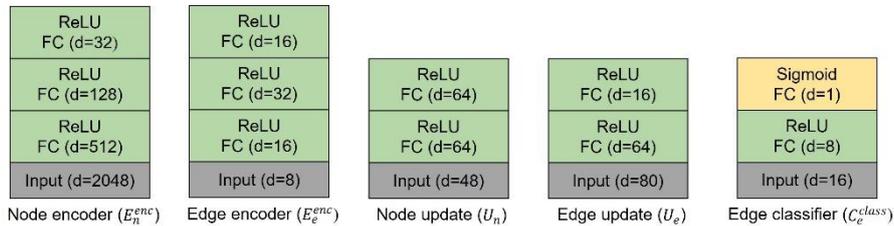

**Fig. 3.** Detailed architectures for all networks. d means the output dimensions for each layer.

**Implementation Details.** We train a YoloX detector to obtain detections for both MOT17 and MOT20 following [1]. The following part describes the implementation details of tracklet association in RTAT. We train a ReID model using ResNet50 following [21] to extract appearance feature. After the convolutional layers in ResNet50, a node encoder is added to reduce the dimension of node feature to 32. All of the networks are light-weight MLPs, and their detailed architectures are shown in Fig. 3.

We use all the tracklets in a video sequence and perform data augmentation from both video-level and tracklet-level to train the GNN networks, as explained in Section 3.4. We use three levels of hierarchical graphs for tracklet association. GNNs in all the three levels are jointly trained for 500 epochs using the Adam optimizer with a learning rate of $3 \cdot 10^{-4}$ and a weight decay of $10^{-4}$. We set $\gamma = 1$ for focal loss, $K$=10 to limit the number of edges connected to each node, and $L$=12 for the steps of message-passing in all GNNs. Furthermore, linear interpolation is applied to fix the missing detections in the final trajectories.

### 4.2 Ablation Studies

We perform 3-fold cross-validation on the MOT17 training set for ablation studies following the experimental setup in [21] and use IDF1 as the primary metric. We study three main aspects of our method in this section. 1) How to select the proper cost threshold $th_c$ to obtain tracklets with high purity. 2) How different training strategies affect the tracklet-association performance. 3) The effect of using different data association methods to generate tracklets in the first-stage.

**Obtain tracklets with high purity.** We select BoT-SORT-ReID [4] to generate tracklets for analysis in this experiment. We set a cost threshold $th_c$ to reject the matchings that have higher cost in the assignment process. Obviously, lower cost threshold can obtain more tracklets with higher purity. However, it is meaningless if we set a very small threshold, such as 0, where each detection is a tracklet with 100% purity. We need to keep a balance between the number of tracklets and the high purity rate (HPR). We list the tracking performance in the first-stage and second-stage with different cost threshold in Table 1. The number of tracklets in the ground-truth is also list for comparison. In the first-stage association, the number of tracklets and HPR constantly increase as we decrease the matching cost threshold, while the IDF1 score decreases slightly when $th_c \geq 0.2$. We perform tracklet association based on the tracklets generated in the first-stage, the IDF1 score has increased after the second-stage association under all cost thresholds, even when $th_c = 0.7$ which is the default setting in BoT-SORT [4]. The best result is achieved when $th_c = 0.2$, which has the highest IDF1 score and the fewest ID switches. When $th_c = 0.1$, the largest increasement on IDF1 (i.e., 19.2%) occurs, however its IDF1 score is lower than that of $th_c = 0.2$ in both stages, and it has much more tracklet (4964 versus 1693) which will bring more computational cost during inference. Therefore, we choose 0.2 as the cost threshold in our experiments.

**Table 1.** The tracking performance in the first-stage and second-stage with different cost threshold. The number of tracklets in ground-truth is also list for comparison

| | First-stage association | | | Second-stage association | | | GT |
|---|---|---|---|---|---|---|---|
| $th_c$ | IDF1 ↑ | #Tracklets | HPR | IDF1 ↑ | #Tracklets | IDs ↓ | #Tracklets |
| 0.7 | 85.0 | 753 | 84.5 | 86.0 | 641 | 291 | |
| 0.5 | 85.2 | 979 | 87.3 | 86.5 | 630 | 282 | |
| 0.4 | 84.5 | 1,091 | 89.0 | 87.6 | 623 | 278 | 546 |
| 0.3 | 84.3 | 1,218 | 91.1 | 88.2 | 615 | 265 | |
| 0.2 | 83.6 | 1,693 | 93.4 | **88.5** | **612** | **261** | |
| 0.1 | 67.1 | 4,964 | 98.1 | 86.3 | 621 | 276 | |

**The effect of different training strategies.** We adopt BoT-SORT-ReID with a cost threshold $th_c = 0.2$ to generate tracklets in the first-stage association.

Firstly, we evaluate how the number of hierarchical levels (*HL*) in the tracklet association effect the metrics of IDF1, IDs and the number of tracklets. The number of

*HL* varies from 0 to 5, and their tracking metrics are list in Table 2. *HL*=0 means the performance for the tracklets obtained in the first-stage. With the increase of *HL*, the IDF1 constantly increase, the ID switches and the number of tracklets constantly decrease. At the same time, both the increase and the decrease become smaller and smaller. The increasement of IDF1 can be ignored when the number of *HL* is bigger than 3. Furthermore, larger number of hierarchical levels will increase the time and memory costs for the tracklet association. Hence, we set *HL*=3 in our experiments.

Secondly, we evaluate the effect of using different data augmentation strategies in training GNN networks. We take all the tracklets in a video sequence to build graph and train GNNs, where a training sample consists of a video sequence and a set of tracklets. However, there are only 7 and 4 training samples in MOT17 and MOT20 respectively, the training samples are very few to learn GNNs with higher robustness and generality. As described in section 3.4, we design data augmentations from both video-level and tracklet-level to generate more training samples. The results of using different combinations of data augmentations are listed in Table 3. We can see that both the data augmentation strategies can improve the IDF1 score separately, and their combination can obtain higher improvement. The results demonstrate the effectiveness of our data augmentation methods in training robust GNN networks.

Table 2. The performance of tracklet association with different hierarchical levels

| # *HL* | IDF1 ↑ | IDs ↓ | # Tracklets |
|---|---|---|---|
| 0 | 83.6 | 1,344 | 1,693 |
| 1 | 86.2 | 406 | 1,021 |
| 2 | 87.6 | 287 | 728 |
| 3 | **88.5** | **261** | **612** |
| 4 | 88.6 | 258 | 608 |
| 5 | 88.6 | 257 | 607 |

Table 3. The performance of tracklet association using different data augmentation

| Data Augmentation | | Tracking Metrics | |
|---|---|---|---|
| Video-level | Tracklet-level | IDF1 ↑ | IDs ↓ |
|  |  | 85.5 | 293 |
| √ |  | 86.7 | 278 |
|  | √ | 87.2 | 272 |
| √ | √ | **88.5** | **261** |

**The effect of using different data association methods in the first-stage.** The aim of first-stage association is to generate tracklets with high purity, which can be achieved by using any trackers with a lower cost threshold. We use three popular trackers, i.e., ByteTrack [1], BoT-SORT [4], Deep OC-SORT[7], for the first-stage association, and compare their performance before and after the tracklet association in second-stage. The results are listed in Table 4. There are big differences among the three trackers on all the three metrics in the first-stage association, however, the differences are largely reduced after the second-stage association. We can see that our method can obtain very similar tracking performance no matter which tracker is used in the first-stage, which indicates that simple data association strategy is good enough for the first-stage. Therefore, our method is helpful to release researchers from the exhausting work of designing more and more sophisticated data association strategy in order to obtain minor improvement in tracking performance.

**Table 4.** The performance of using different data association methods in the first-stage.

|  | First-stage association | | | Second-stage association | | |
| --- | --- | --- | --- | --- | --- | --- |
| Tracker | IDF1 ↑ | #Tracklets | HPR | IDF1 ↑ | #Tracklets | IDs ↓ |
| ByteTrack | 76.6 | 1,571 | 91.9 | 88.0 | 623 | 276 |
| BoT-SORT | 83.6 | 1,693 | 93.4 | **88.5** | **612** | **261** |
| Deep OC-SORT | 81.2 | 1,264 | 89.7 | 88.2 | 617 | 264 |

**Table 5.** Comparison of the state-of-the-art methods under the "private detection" protocol on MOT17 test set. The trackers are sorted by HOTA. The best results are shown in **bold**.

| Tracker | HOTA ↑ | IDF1 ↑ | MOTA ↑ | AssA ↑ | IDs ↓ |
| --- | --- | --- | --- | --- | --- |
| ByteTrack [1] | 63.1 | 77.3 | 80.3 | 62.0 | 2,196 |
| StrongSORT [5] | 64.4 | 79.5 | 79.6 | 64.4 | 1,194 |
| Deep OC-SORT [7] | 64.9 | 80.6 | 79.4 | 65.9 | 1,023 |
| BoT-SORT [4] | 65.0 | 80.2 | 80.5 | 65.5 | 1,212 |
| MotionTrack [8] | 65.1 | 80.1 | 81.1 | 65.1 | 1,140 |
| ConfTrack [36] | 65.4 | 81.2 | 80.0 | 66.3 | 1,155 |
| CBIOU [14] | 66.0 | 82.5 | **82.8** | 66.1 | 1,194 |
| PIA [38] | 66.0 | 81.1 | 82.2 | 65.8 | 1,026 |
| ImprAsso [10] | 66.4 | 82.1 | 82.2 | 66.6 | 924 |
| SUSHI [22] | 66.5 | 83.1 | 81.1 | 67.8 | 1,149 |
| **RTAT-ByteTrack (ours)** | 67.0 | 84.4 | 80.1 | 69.3 | 942 |
| **RTAT-BoT-SORT (ours)** | **67.2** | **84.7** | 80.4 | **69.7** | **912** |

### 4.3 Benchmarks Evaluation

We present the results of the state-of-the-art trackers on the test set of MOT17 and MOT20 benchmarks under the "private detection" protocol in Table 5 and Table 6, respectively. All the results are obtained from the official MOTChallenge server [37]. We adopt ByteTrack and BoT-SORT to generate tracklets in the first-stage association, which are named RTAT-ByteTrack and RTAT-BoT-SORT, respectively. Both versions of our method outperform all the other trackers in almost all the main metrics. Our method can achieve the best performance in all association related metrics, i.e., HOTA, IDF1, and AssA, on both benchmarks, which demonstrate the effectiveness of our method for data association. For example, RTAT-BoT-SORT outperforms the tracker in second place by a large margin (i.e., +1.4 HOTA, +2.3 IDF1, and +1.9 AssA) on MOT20 benchmark.

Both RTAT-ByteTrack and RTAT-BoT-SORT outperform their respective baseline by a large margin on both MOT17 and MOT20. It is worth noting that RTAT-

ByteTrack can achieve similar performance with RTAT-BoT-SORT in all metrics. The performance gap between ByteTrack and BoT-SORT are filled by using the tracklet associations in our method. This observation demonstrates that simple association strategy is enough to generate tracklets with high purity for the tracklet association in the second-stage, and there is no need to design more sophisticated data association strategy by investing a great deal of time and effort.

Table 6. Comparison of the state-of-the-art methods under the "private detection" protocol on MOT20 test set. The trackers are sorted by HOTA. The best results are shown in **bold**.

| Tracker | HOTA↑ | IDF1↑ | MOTA↑ | AssA↑ | IDs↓ |
| --- | --- | --- | --- | --- | --- |
| ByteTrack [1] | 61.3 | 75.2 | 77.8 | 59.6 | 1,223 |
| StrongSORT [5] | 62.6 | 77.0 | 73.8 | 64.0 | 770 |
| MotionTrack [8] | 62.8 | 76.5 | 78.0 | 61.8 | 1,165 |
| BoT-SORT [4] | 63.3 | 77.5 | 77.8 | 62.9 | 1,313 |
| FineTrack [9] | 63.6 | 79.0 | 77.9 | 63.8 | 980 |
| Deep OC-SORT [7] | 63.9 | 79.2 | 75.6 | 65.7 | 779 |
| SUSHI [22] | 64.3 | 79.8 | 74.3 | 67.5 | 706 |
| ImprAsso [10] | 64.6 | 78.8 | **78.6** | 64.6 | 992 |
| PIA [38] | 64.7 | 79.0 | 78.5 | 64.9 | 1,023 |
| ConfTrack [36] | 64.8 | 80.2 | 77.2 | 66.2 | **702** |
| **RTAT-ByteTrack (ours)** | 65.9 | 82.1 | 78.1 | 67.7 | 817 |
| **RTAT-BoT-SORT (ours)** | **66.2** | **82.5** | 78.4 | **68.2** | 787 |

## 5    Conclusion

We propose a Robust Two-stage Association Tracker (RTAT), which can achieve higher association performance by utilizing the advantages of two kinds of data association methods: the simplicity and efficiency of handcrafted association methods and the effective high-order contextual information of learnt association methods. We use a simple data association method to generate tracklets with high purity in the first-stage and use message-passing GNNs to perform tracklet association in the second-stage. We further design data augmentation strategies from video-level and tracklet-level to improve the generalization ability of our tracklet association model. Ablation studies and MOT benchmarks results validate the effectiveness of our method. We hope our work is helpful to release researchers from the hard and exhausting work of designing more and more sophisticated data association strategy to obtain minor improvement in tracking performance. We also expect this work can push forward the development of multiple-object tracking.